\documentclass{article}

\usepackage[preprint]{neurips_2025}

\usepackage[utf8]{inputenc} 
\usepackage[T1]{fontenc}    
\usepackage{hyperref}       
\usepackage{url}            
\usepackage{booktabs}       
\usepackage{amsfonts}       
\usepackage{nicefrac}       
\usepackage{microtype}      
\usepackage{xcolor}         
\usepackage{graphicx}  
\usepackage{subcaption}
\usepackage{amsmath}
\usepackage{wrapfig}

\title{Inferring Effects of Major Events through \Task~of Population Anxiety}

\author{%
  Siddharth Mangalik \\
  Department of Computer Science\\
  Stony Brook University\\
  \texttt{smangalik@cs.stonybrook.edu} \\
  \And
  Ojas Deshpande \\
  Department of Computer Science\\
  Stony Brook University \\
  \texttt{odeshpande@cs.stonybrook.edu} \\
  \And
  Adithya V. Ganesan \\
  Stony Brook University \\
  Department of Computer Science\\
  \texttt{avrinchipurum@cs.stonybrook.edu} \\
  \And
  Sean A. P. Clouston \\
  Department of Family, Population, and Preventive Medicine \\
  Renaissance School of Medicine at Stony Brook University \\
  \texttt{sean.clouston@stonybrookmedicine.edu} \\
  \And
  H. Andrew Schwartz \\
  Department of Computer Science\\
  Stony Brook University \\
  \texttt{has@cs.stonybrook.edu} \\
}

\newcommand{\Task}{\text{Discontinuity Forecasting}}
\newcommand{\task}{\textit{discontinuity forecasting}}
\newcommand{\countyEmbeddingModel}{\text{RoBERTa-Large}}

\begin{document}

\maketitle



\begin{abstract} 
Estimating community-specific mental health effects of local events is vital for public health policy. 
While forecasting mental health scores alone offers limited insights into the impact of events on community well-being, quasi-experimental designs like the Longitudinal Regression Discontinuity Design (LRDD) from econometrics help researchers derive more effects that are more likely to be causal from observational data. 
LRDDs aim to extrapolate the size of changes in an outcome (e.g. a discontinuity in running scores for anxiety) due to a time-specific event. 
Here, we propose adapting LRDDs beyond traditional forecasting into a statistical learning framework whereby future discontinuities (i.e. time-specific shifts) and changes in slope (i.e. linear trajectories) are estimated given a location's history of the score, dynamic covariates (other running assessments), and exogenous variables (static representations). 
Applying our framework to predict discontinuities in the anxiety of US counties from COVID-19 events, we found the task was difficult but more achievable as the sophistication of models was increased, with the best results coming from integrating exogenous and dynamic covariates. 
Our approach shows strong improvement ($r=+.46$ for discontinuity and $r = +.65$ for slope) over traditional static community representations. 
Discontinuity forecasting raises new possibilities for estimating the idiosyncratic effects of potential future or hypothetical events on specific communities. 
\end{abstract}

\section{Introduction}

In 2019, a novel coronavirus disease (COVID-19) emerged and caused a global pandemic~\citep{CovidW2020}. The global response was varying, with some communities unable to respond while others implemented a varying degree of public health responses ~\citep{deng2020characteristics,basseal2023key,scally2020uk}. Such responses included social distancing, shifting workplaces or schools outdoors, protective masking, improving air circulation and filtering processes, and school/workplace closures. 

As COVID-19 began circulating, studies reported 27.6\% and 25.6\% increases in the prevalence of major depression and anxiety disorders worldwide~\citep{santomauro2021global}. Population mental health is sensitive to changes in social context~\citep{boden2021addressing}, so while the source of poorer mental health was discussed, many researchers positing that COVID-19 shutdowns are to blame for poorer mental health~\citep{tull2020psychological, hossain2020epidemiology, kumar2021covid}. Notably, resaerchers have pointed to poo part due to a broad impact of closures on mental health in children ~\citep{mazrekaj2024impact, viner2022school} and vulnerable adults ~\citep{twenge2020mental, sahu2020closure}. Amazingly, this story of mental health was dominant during and after the pandemic so that reviews often either ignore the impact of the arrival of COVID-19 on population mental health or characterize the impact as one of many potential causes of mental health alongside the closures~\citep{talevi2020mental,xiong2020impact,vindegaard2020covid}. One result of this lack of focus may be that uture pandemic responses may consider mental health consequences of a pandemic as a counterbalancing force against the need for a robust public health response. However, this may be due to the fact that no studies have examined whether the infiltration of COVID-19 into individual communities is a source of anxiety or depression. 

COVID-19 taught us that the ability to understand the effect of size and multidimensional causes of changes in population mental health were not only important as outcomes but that they were also important for understanding the public's support for public health efforts on the short term, and critical for maintaining ongoing support over the long term. The ability to forecast changes in community health is vital to both understanding the effects of health-related community events as well as providing differentiated and targeted responses ~\citep{rose2001sick, Nsubuga2006, luhmann2023loneliness}. Yet, at the population level, anxiety and depression are likely to be associated with markers of socioeconomic status (SES)~\citep{everson2002epidemiologic} that are also commonly linked to a greater vulnerability to the worst impacts of COVID-19 ~\citep{clouston2021socioeconomic}.  
Individual-level risk factors for anxiety and depression are increasingly well understood, but community-level effects beyond socio-economics have been more elusive.  

One reason for a limited understanding of the impact of COVID-19 on population mental health may be that data sources used targeted population-based approaches. These approaches focus on specific vulnerable populations like school children, older individuals, or inpatients with COVID-19 to understand mental health. These targeted studies, though, lack temporal and geographic variability and, therefore, lack the ability to specifically examine how COVID-19 might increase anxiety while the COVID-19 closures might also cause depression in those who are most affected. Recent work at the intersection of epidemiology, machine learning, and psychology has proposed to monitor mental health by generating robust and generalizable language-based weekly U.S. county-level measurements of anxiety and depression, reaching greater reliability at finer time and space resolutions than survey-based methods~\citep{mangalik2024robust}. These data now provide us with the temporal and geographic variability necessary to use "longitudinal discontinuity forecasting": a causal modeling technique that seeks to disentangle the true causal effects of major events, like the COVID-19 pandemic, from other temporally-variant confounding influences that could obscure underlying causal effects ~\citep{thistlethwaite1960regression, hahn2001identification}. 

In this study, we utilize temporally and geographically varying nationally representative mental health assessments alongside supervised learning techniques employed for time-series analysis to predict community-specific discontinuities from major events -- \task. 
Compared to typical forecasting, \task~is a more principled approach for modeling changes in groups after major shocks, similar though not identical to randomized control trials for estimating treatment effects~\citep{maas2017regression}. 
Our approach centers on the use of causal inference methods, enhanced with endogenous timeline variables as well as exogenous and static variables, which have found use in time-series forecasting with Transformers~\citep{wang2024timexer, shang2024ada}. The use of such a system could enable the prediction of the impact of events on communities that are poised for particular incidents such as disease outbreak, natural disasters, or violence.

\section{Contributions}
The main contributions of this work include:
(1) A proposed regression discontinuity method to assess the effect of community-specific events on health assessments. 
(2) The evaluation of the effect of COVID-19 incidence and death on real-world county anxiety assessments.
(3) Demonstrating the ability of supervised learning techniques to predict how idiosyncratic communities will respond to a given event using \textit{endogenous} variables (timelines and regression coefficients) and \textit{exogenous} variables (language embeddings and time-aware covariates).
(4) Investigation of how much \task~can be attributed to community sociodemographic and urbanicity differences.
To the best of our knowledge, this is the first investigation of the quasi-experimental technique, Longitudinal Regression Discontinuity Design, as a prediction (i.e. statistical learning with out-of-sample testing) task.

\section{Problem Formulation}

\begin{figure}
    \centering
    \includegraphics[width=.495\linewidth]{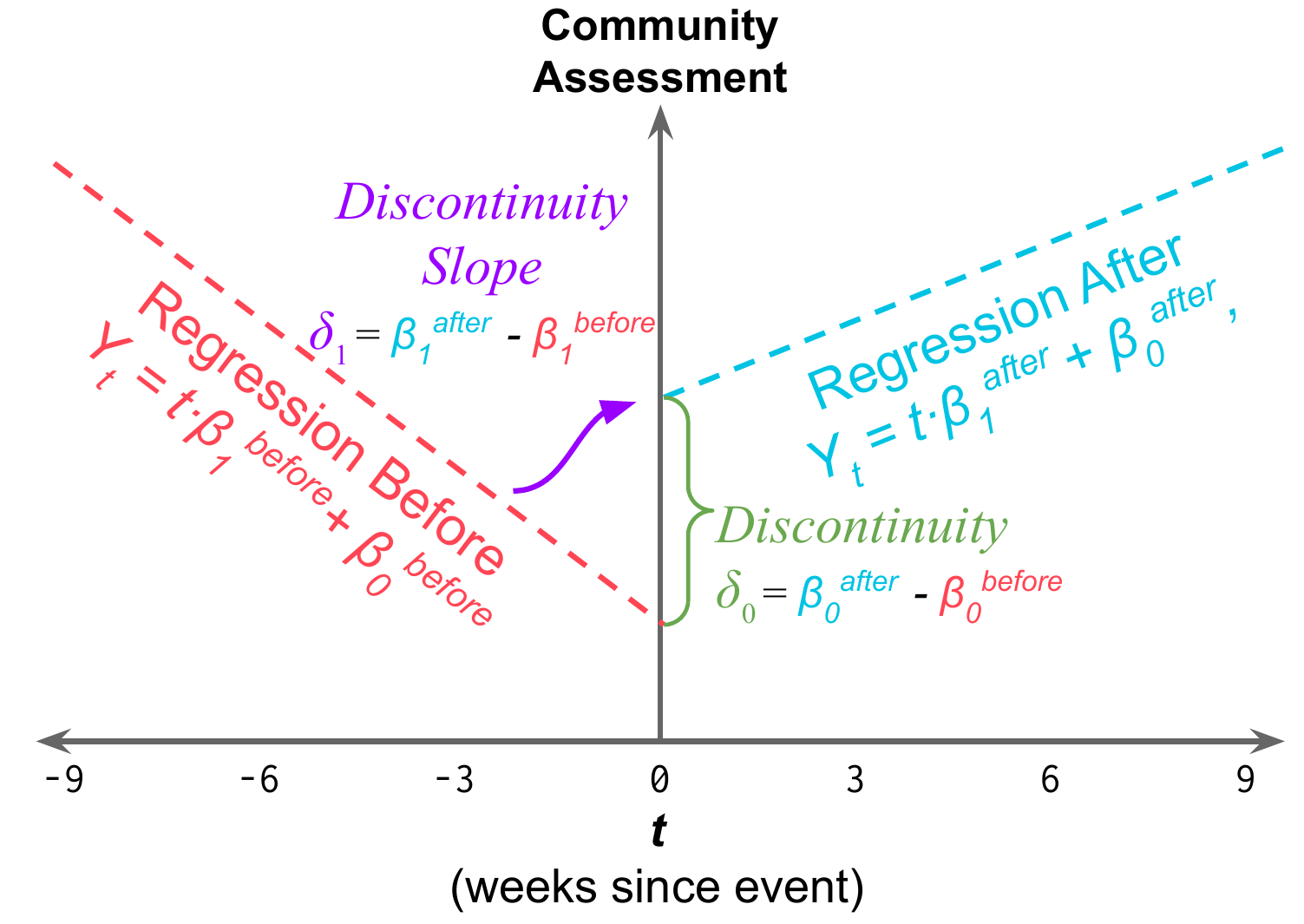}
    \caption{Calculation of discontinuity ($\delta_0$, green) and discontinuity slope ($\delta_1$, purple) before and after a critical event at $t=0$, using regression parameters for the community assessment before (red) and after (teal) the event.}
    \label{fig:explanatory_figure}
\end{figure}

Regression Discontinuity Designs (RDDs) find evidence for causal effects by comparing an outcome of interest, just before and after a determined cutoff in a continuous variable~\citep{thistlethwaite1960regression, hahn2001identification} (see Figure \ref{fig:explanatory_figure}).
The assumption is that any discontinuity around the cutoff is due to the critical event threshold, with the canonical example being an event of receiving a scholarship at an SAT score of 1200 and observing a discontinuity (an improvement in this case) in college grade point average. 
Such methods have seen use in econometrics, political and social sciences, and epidemiology as a canonical method for supporting causal claims on empirical data~\citep{leamer1983let}. 
RDDs belong to a class of quasi-experimental designs \citep{harris2006use, angrist2009mostly, liu2021quantifying} that approximate ``gold-standard'' experiments where subjects are randomly assigned to either a control or treatment group.
Such designs are vital for studying situations where it is too expensive, practically infeasible, or even unethical to randomly assign treatments to different populations~\citep{musci2019ensuring}.



Formally, an RDD estimate of change is the difference between the mean findings when approaching the event from the right (after) and the mean approaching from the left (before)~\citep{imbens2008regression, lee2010regression}.
For each series (e.g. anxiety in a county) one observes a running variable $Y_t$ (e.g.\ weekly anxiety) over a window of $2T+1$ time points centered on an event at $t=0$, i.e.\ $Y_{-T},\dots, Y_0,\dots, Y_{T}$ (here $T=9$).  The discontinuity ($\delta_0$) and discontinuity slope ($\delta_1$) then describe the effect of the event at $t=0$, as shown in \autoref{fig:explanatory_figure}:
\[
\begin{aligned}
&\text{Discontinuity:}\quad \delta_0 \;=\;\lim_{t\to0^+} E[Y_t] \;-\;\lim_{t\to0^-} E[Y_t]\;,\\
&\text{Discontinuity Slope:}\quad \delta_1 \;=\;\lim_{t\to0^+} \frac{d}{dt}E[Y_t] \;-\;\lim_{t\to0^-} \frac{d}{dt}E[Y_t]\;.
\end{aligned}
\]
In practice, $\delta_0$ and $\delta_1$ are estimated using linear regressions fit on the ``before'' ($t=-T,\dots,-1$) and ``after'' ($t=1,\dots, T$) windows:
\[
\begin{aligned}
\hat\beta_1^{(\text{before})}
&=\frac{\sum_{t=-T}^{-1}(t-\bar t)(Y_t-\bar Y)}{\sum_{t=-T}^{-1}(t-\bar t)^2},\quad
\hat\beta_0^{(\text{before})} = \bar Y - \hat\beta_1^{(\text{before})}\,\bar t.\\
\end{aligned}
\]
The above is repeated for the ``after'' window, resulting in the outcomes:
\[
\begin{aligned}
\delta_1 &= \hat\beta_1^{(\text{after})} - \hat\beta_1^{(\text{before})},\quad
\delta_0 = \hat\beta_0^{(\text{after})} - \hat\beta_0^{(\text{before})}.
\end{aligned}
\]

\paragraph{\Task~as a Predictive Task.}
We introduce \textbf{\task}, the task of predicting a future discontinuity (both $\delta_0$ and $\delta_1$) in an outcome ($Y$) at a specified event time $t=0$ given a history of the endogenous outcome preceding the event ($y_{-T:t-1}$), that history's regression coefficients ($\beta_{0,-T:t-1}, \beta_{1,-T:t-1}$), two types of covariates preceding the event ($X_{exog}$, $X_{cov, -T:t-1}$), and no post-event data:  
\[
(\hat{\delta}_0,\hat{\delta}_1) \;=\; f(y_{-T:t-1}, \beta_{0,-T:t-1}, \beta_{1,-T:t-1} ,X_{exog}, X_{cov, -T:t-1}) 
\]
where $X_{exog}$ represents \textit{static exogenous descriptors} of the sequence (e.g. demographic attributes of a location) while $X_{cov}$ represents \textit{dynamic covariates} that can change over time alongside the endogenous outcome (e.g. an outcome of anxiety scores using depression as dynamic covariates). 




We then treat \task~as a multi-output regression problem\footnote{While we use a MSE multi-task objective with equal weighting, any multi-task regression loss can be used.}:
\[
\min_{f}\; \mathbb E\bigl[(\hat\delta_0 - \delta_0)^2, (\hat\delta_1 - \delta_1)^2\bigr].
\]
where $f$, $\hat\delta_0$, and $\hat\delta_1$ represent the model, estimates of discontinuity and discontinuity slope, respectively. Gold standard values for $\delta_0$ and $\delta_1$ are derived from the LBMHA dataset.


\begin{figure}[!t]
    \centering
    \begin{subfigure}[t]{.495\linewidth}
        \centering
        \includegraphics[width=\linewidth]{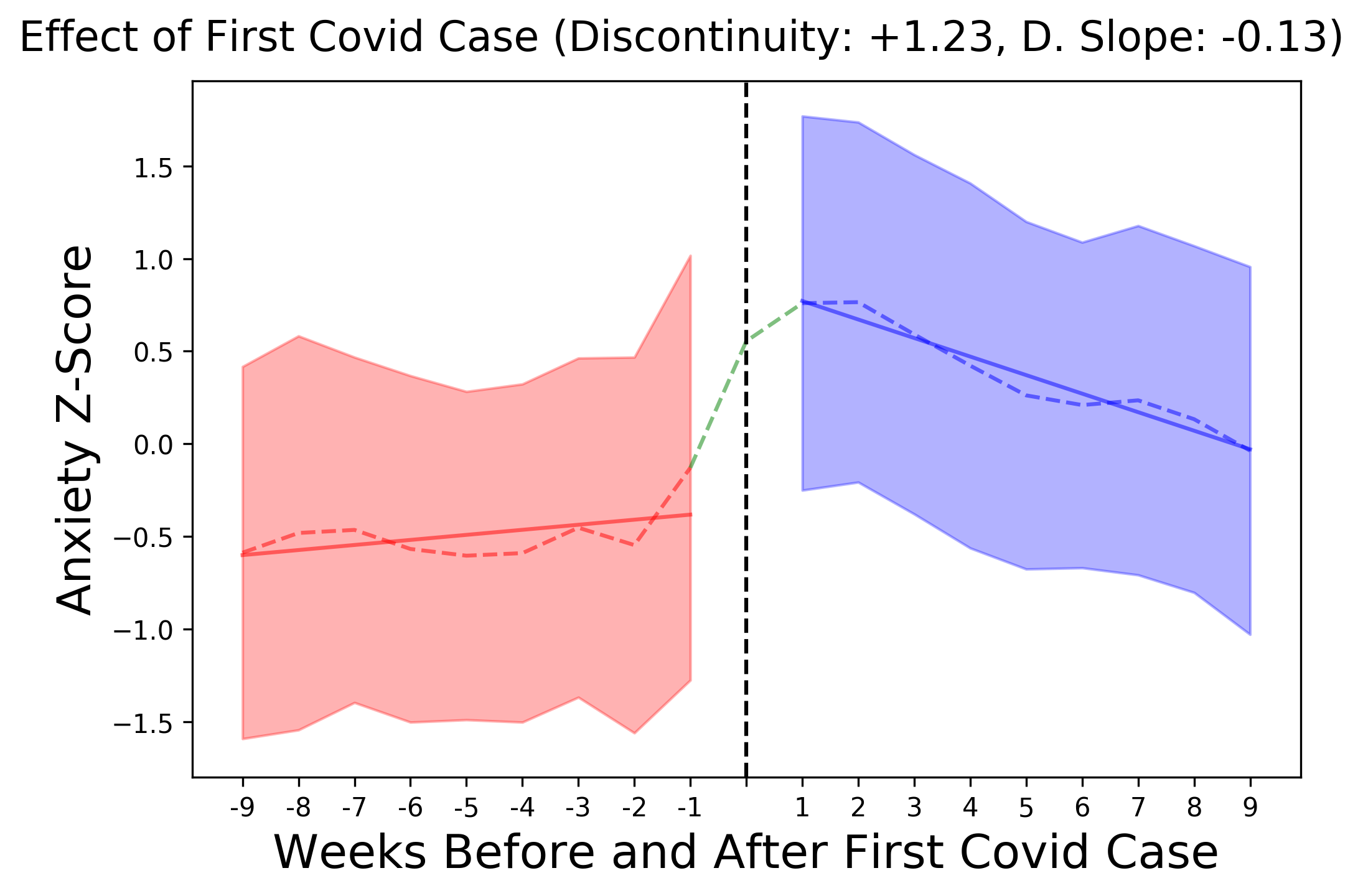}
        \caption{}
        \label{fig:rdd_spaghetti_covid_case}
    \end{subfigure}
    \begin{subfigure}[t]{.495\linewidth}
        \centering
        \includegraphics[width=.98\linewidth]{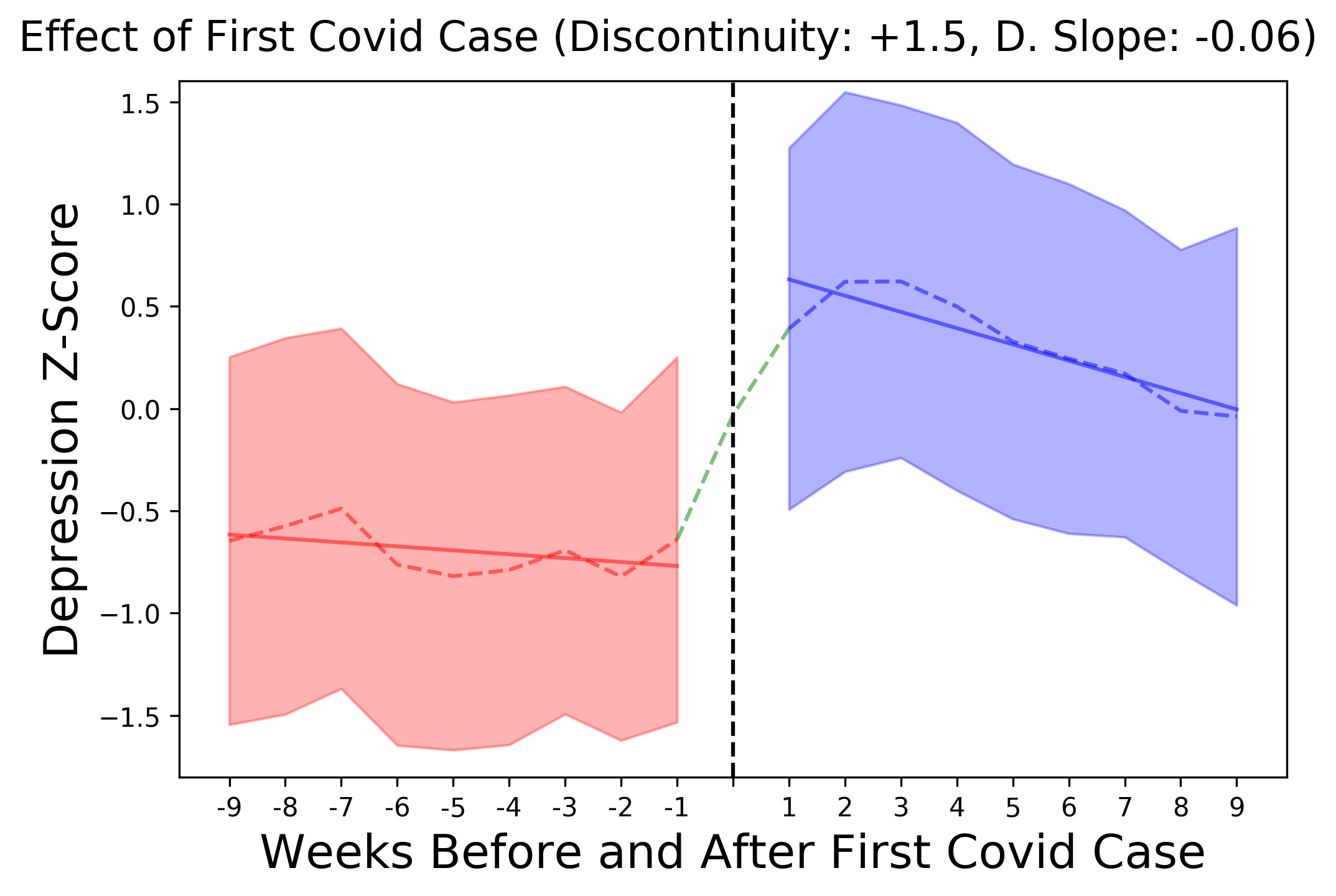}
        \caption{}
        \label{fig:rdd_spaghetti_covid_case_depression}
    \end{subfigure}
    \begin{subfigure}[t]{.495\linewidth}
        \centering
        \includegraphics[width=\linewidth]{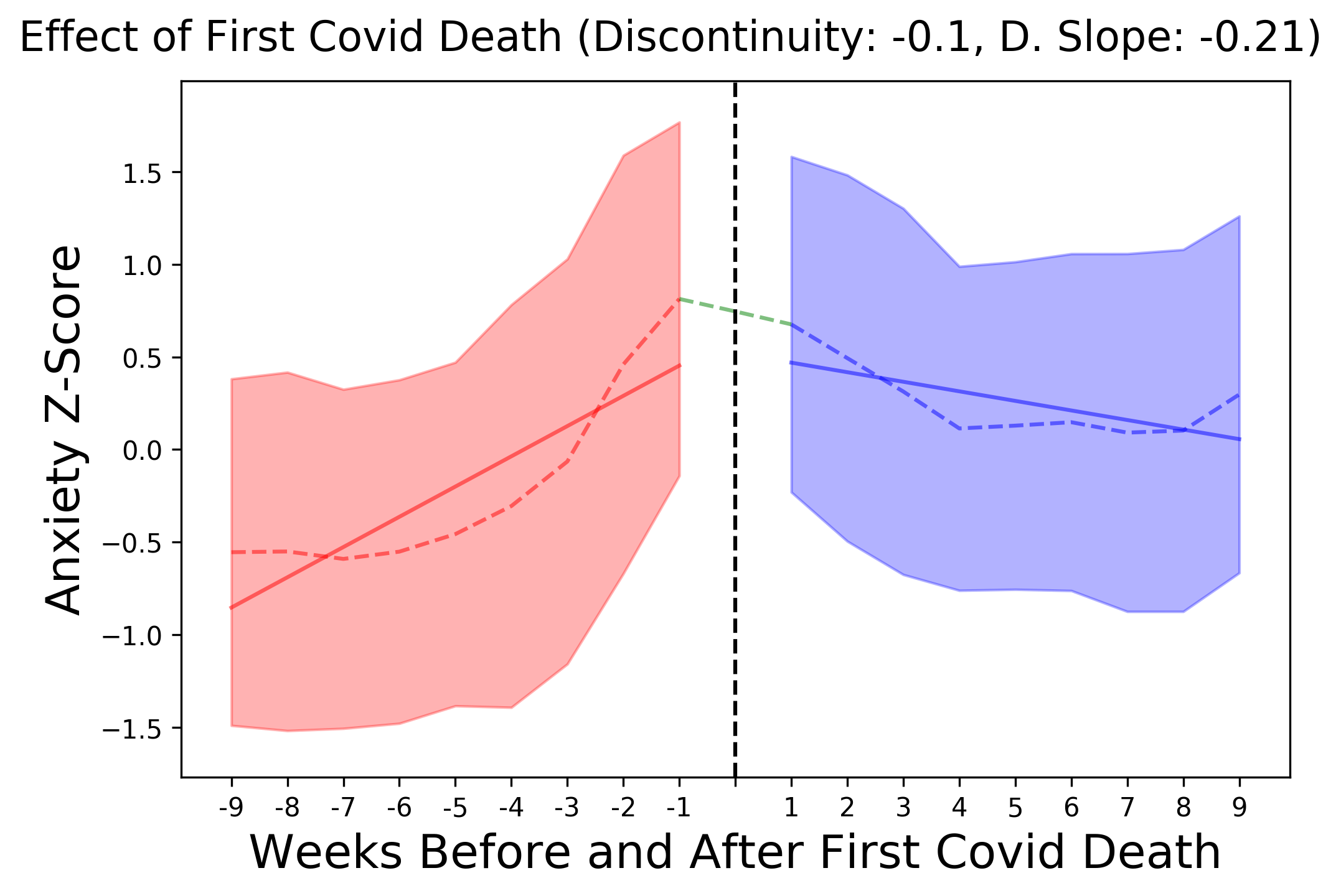}
        \caption{}
        \label{fig:rdd_spaghetti_covid_death}
    \end{subfigure}
    \begin{subfigure}[t]{.495\linewidth}
        \centering
        \includegraphics[width=.965\linewidth]{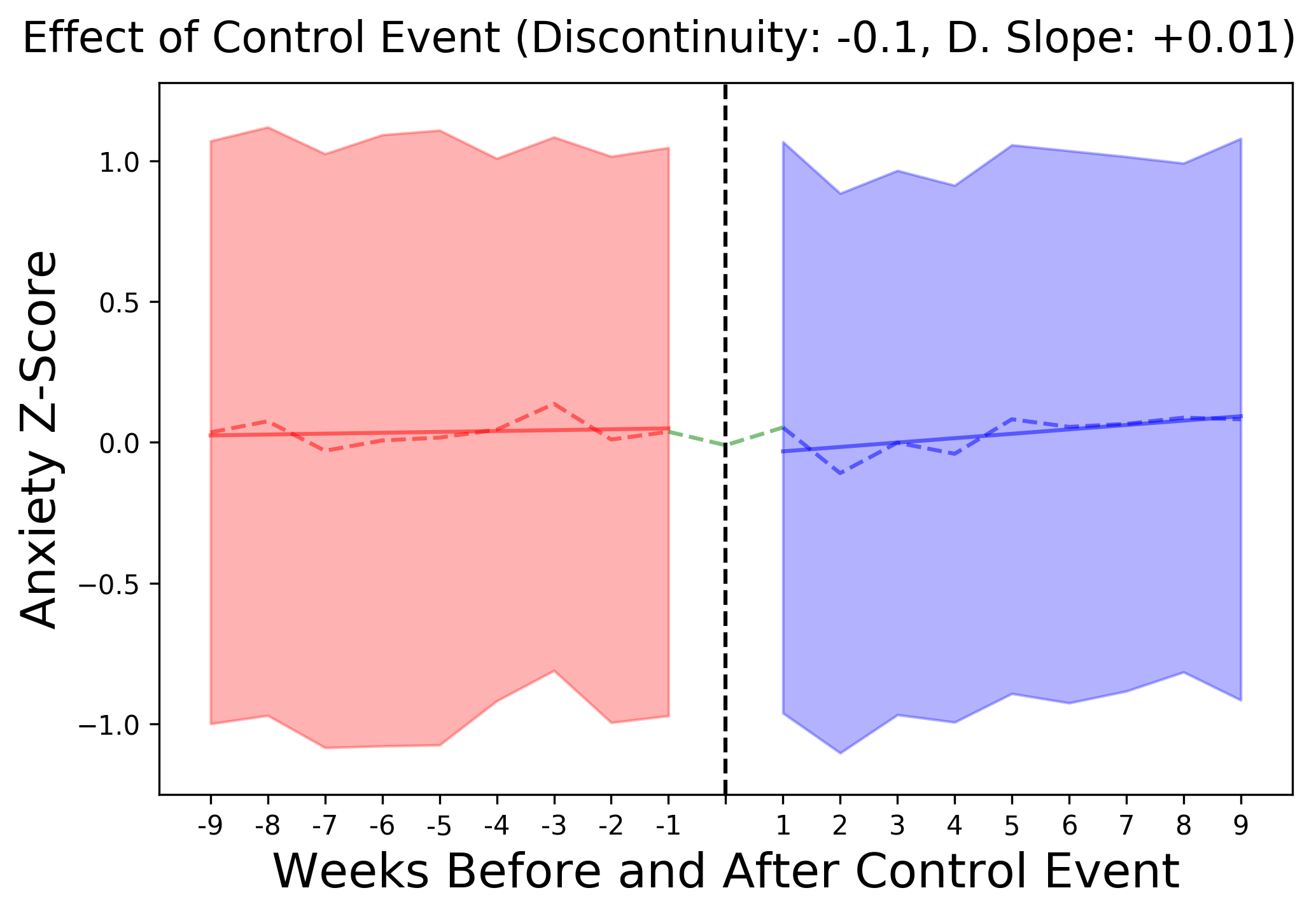}
        \caption{}
        \label{fig:rdd_spaghetti_control}
    \end{subfigure}
    \caption{Mean anxiety (a) and depression (b) before and after the first case of COVID-19 in US counties. Measuring the anxiety change from the first death from COVID-19 in a US county (c), and running the same analysis against a randomized control event (d). The gap in assessments reflect the 1-week buffer we apply to pre- and post-event periods. Dashed lines are the weekly mean mental health score, based on offset from the occurrence of the event (vertical black line). Shading represents one standard deviation in the mean measurement.}
    \label{fig:rdd_spaghetti}
\end{figure}

\section{Experiments}

\subsection{Language-Based Mental Health Assessments}
For our running variable, we use county-week language-based assessments of anxiety from the LBMHAs dataset~\citep{mangalik2024robust}. 
This dataset derives these assessments from $\sim$1 billion English tweets made in the US from 2019 to 2020 after weighting Twitter users to better represent the communities they live in. The findings in this work were found to hold convergent validity with a comparable survey, alongside greater external validity than the survey against measures of community politics, economics, social life, and health.  

We track the weekly anxiety assessments of US counties that exhibited high reliability, as measured through the number of unique users recorded within a given county and week. 
We use the ``differenced'' data for our analyses, which controls for seasonal influences on the mental health assessments, and filter county-week measurements to those containing measurements from at least 200 unique individuals, which is the documented threshold for reliable assessment. 
We then Z-score the per-county findings such that the original scores, which were scaled from 0-5, are now county mean-centered at 0. 

We compare this language-based data to contemporary survey-based findings, the best-in-class being Gallup's COVID Panel Data from the same period in 2020~\citep{gallup2021covid}. This survey data was collected shortly after the onset of COVID-19 to collect weekly self-reports of worry and sadness across US counties and contained on average 4,601 measurements/week for 22 weeks, covering 2,617 counties. When we attempted to run our analyses against this dataset, we found that while it broadly covered the US it only reported an average of 8.75 measurements per county-week, well below the 50 measurements found to be necessary for reliability~\citep{mangalik2024robust}. 
This sparseness made it impossible to conduct Longitudinal RDDs with survey data and also highlighted that, since the data was collected in response to COVID-19, there was no retrospective data collected before March 2020 that could be used to measure mental health effects from COVID-19 events. LBMHAs by comparison provide significantly more coverage in time and space alongside their established reliability and external validity.

\subsection{COVID-19 Incidence and Deaths}
County event data was collected from The New York Times, based on reports from state and local health agencies. 
Specifically, we use their Cumulative Cases and Deaths for COVID-19 dataset~\citep{nytimes2023}, which contains the number of cases and deaths for all counties in the U.S. every day from January 21, 2020 (the first recorded case in the United States) until April 2022. 
For all counties we study the data includes both a first hospitalization and a first death date.

When analyzing major events, we account for the possibility that, although the event has occurred within an individual's community, they may not have been aware of it. We also consider the case of counties expressing anxiety before the event reaches their community from viewing national news.
From a counterfactual perspective, the news that COVID-19 is incoming but not present in one's community is important in the timing of anxiety changes. 
To more clearly capture a causal inference of our events, we apply a one-week buffer around the event to form pre- and post-envelope periods, and then difference those periods. 
\autoref{fig:rdd_spaghetti} demonstrates this buffering, while \autoref{sec:ablations} shows the original unbuffered timeline and the results from extending the buffer.

\begin{table}[!t]
\centering
\begin{tabular}{lcccc} \toprule
 & \multicolumn{2}{c}{\textbf{Discontinuity}} & \multicolumn{2}{c}{\textbf{Discontinuity Slope}} \\ \cmidrule(lr){2-3}\cmidrule(lr){4-5} 
 & MSE ($\downarrow$) & r ($\uparrow$) & MSE ($\downarrow$) & r ($\uparrow$) \\ \midrule
No Change Baseline & 2.239 & - & .126 & - \\
Mean Baseline & 1.222 & - & .124 & - \\ 
Forecasting Baseline & 1.791 & .582 & .059 & .770 \\ \midrule
\multicolumn{5}{l}{\textbf{GOFAI:}} \\ \midrule
Ridge & .620 & .709$^\dagger$ & .031 & .865$^\dagger$ \\
kNN & .550 & .760$^\dagger$ & .033 & .862$^\dagger$ \\
FFN & .547 & .745$^\dagger$ & .039 & .832$^\dagger$ \\ \midrule
\multicolumn{5}{l}{\textbf{Ensembles:}} \\ \midrule
Random Forest & .530 & .754$^\dagger$ & \textbf{.031} & \textbf{.872}$^\dagger$ \\
Extra Trees & \textbf{.521} & \textbf{.760}$^\dagger$ & .031 & .861$^\dagger$ \\
XGBoost & .773 & .647$^{**}$ & .044 & .865$^{**}$ \\ \bottomrule \\
\end{tabular}
\caption{Establishing the best non-sequential modeling \task~results against baselines by evaluating the models on a held out set of LBMHA outcomes. Mean squared errors and Pearson correlations are reported for predicting discontinuities and discontinuity slopes of anxiety scores after a county has its first case of COVID-19. Paired t-test statistical significance to $p\leq.05^*$, $.01^{**}$, $.001^\dagger$ performance over the mean baseline}
\label{tab:model_p_rc}
\end{table}

\subsection{Predictive Models}
For \task~training, we compare across model families: 
\emph{GOFAI}: Linear ridge regression, $k$-Nearest Neighbors (kNN), and feed-forward networks (FFN).
\emph{Ensembles}: Random Forest, Extra Trees, XGBoost.
And \emph{Sequential NNs}: GRUs and Transformers.

All models were trained with a fixed random seed on identical sampled county timelines from 60\% (217) of the counties, with 20\% (72) of the counties being reserved as a dev set for hyperparameter tuning, and 20\% (72) of the counties held out for model evaluation reporting (See \autoref{sec:hyperparameters} for the exact hyperparameters used). Mean-squared error (MSE) was used to calculate the loss, and then both MSE and Pearson $r$ were reported as performance metrics against the test counties. 

Our GOFAI models all outperform the baseline models (more details in \autoref{sec:baselines}); these findings were all statistically significant.
Ensemble models, in particular Random Forest and Extra Trees, were well suited for this task and performed better than a more complex feed-forward network.
Our results are reported in \autoref{tab:model_p_rc}.  

We examine the incremental value of training with additional exogenous county representations and time-aware covariates (See \autoref{tab:picking_features} for a direct comparison of these feature sets):
\[
\text{(endogenous only)} \;\to\; (+\!\text{dynamic covariates}) \;\to\; (+\!\text{exogenous embeddings}) \;\to\; (\text{all combined})
\]
\paragraph{Dynamic Covariates}
Alongside our anxiety assessment scores, we have corresponding depression timelines, as depression has shown high comorbidity with anxiety~\citep{gorman1996comorbid, pollack2005comorbid,coplan2015treating}. 
We use these scores along with their own respective fitted linear coefficients as dynamic covariates, and we include a single stream to not over-index on covariate motion~\citep{trippe2021high}. 
We see greater improvement from using county embeddings over the dynamic covariates, but for tree-based models, we find that they degrade performance, likely because smaller models are not capabale of handling the higher cardinality of the feature sets. Findings after including these variables can be found in \autoref{tab:model_p_rc_cov_e}.

\paragraph{Exogenous Embeddings}
Following on previous work that sought to generate region-specific embeddings~\citep{hui2020predicting, bevara2024census2vec}, we create 1,422 county embeddings by first extracting the second-to-last layer \countyEmbeddingModel~embedding against every tweet in the County Tweet Lexical Bank~\citep{giorgi2018remarkable, mangalik2024robust} from 2019 to 2020. 
This 1,024-length tweet embedding is then mean-aggregated to Twitter (now X) users before being mean-aggregated again to the county that the user resides in. 
Aggregating through users has been found to yield improvements in downstream modeling performance~\citep{giorgi-etal-2018-remarkable}.
Our approach shows strong improvement ($r=+.46$ for discontinuity and $r = +.65$ for slope) over traditional static community representations.
We report results for the use of these features in \autoref{tab:model_p_rc_cov_e}. 

\autoref{tab:model_seq} demonstrates our evaluations against larger contextual models that are more capable of synthesizing large amounts of inputs. 
We find that these large models performed best when provided with the full set of external variables.
Notably, ensemble models were adequate for this task with only endogenous variables and performed comparably to sequential contextual models using considerably more features.


\begin{table*}[!t]
\centering
\begin{tabular}{lcccc} \toprule
 & \multicolumn{2}{c}{\textbf{Discontinuity}} & \multicolumn{2}{c}{\textbf{Discontinuity Slope}} \\ \cmidrule(lr){2-3}\cmidrule(lr){4-5} 
 & MSE ($\downarrow$) & r ($\uparrow$) & MSE ($\downarrow$) & r ($\uparrow$) \\ \midrule
Ridge                   & .620 & .709$^\dagger$ & \textbf{.031} & .865$^\dagger$ \\
Ridge (+cov)            & .604 & .714 & .033 & \textbf{.867} \\ 
Ridge (+exog)           & .619 & .709$^*$ & .031 & .865 \\ 
Ridge (+exog) (+cov)    & .603 & .714 & .033 & .867 \\ \midrule
kNN                 & .550 & .760 & .033 & .862 \\
kNN (+cov)          & .678 & .751 & .034 & .857 \\
kNN (+exog)         & .542 & \textbf{.761}$^*$ & .035 & .852 \\
kNN (+exog) (+cov)  & .704 & .674 & .035 & .866 \\
\midrule
Extra Trees                 & \textbf{.521} & .760 & .031 & .861 \\
Extra Trees (+cov)          & .543 & .744 & .034 & .855 \\ 
Extra Trees (+exog)         & .568 & .734 & .061 & .746 \\ 
Extra Trees (+exog) (+cov)  & .562 & .736 & .056 & .783 \\ 
\bottomrule
\end{tabular}
\caption{Comparing the effect of including exogenous (+exog) and dynamic covariates (+cov) for \task. The exogenous variables are county embedding inputs and the dynamic covariates (+cov) are depression assessments over time. Significance testing was done on whether the added features improved the model over the base features.}
\label{tab:model_p_rc_cov_e}
\end{table*}

\subsection{LRDD Validity}
To validate and create a baseline for our implementation of LRDDs, we evaluate the same anxiety assessments against a ``control'' event where a random event date is selected for every county. \autoref{fig:rdd_spaghetti_control} visually depicts a single random set of events for this experiment, compared to a Discontinuity for the event ``First Case of COVID''. Repeating this event randomization process over 5,000 county timelines resulted in a mean anxiety Z-score Discontinuity of -.024 (standard deviation = .097) and a mean Discontinuity Slope of .001 (standard deviation = .042) for our control event.

\subsection{Individual County Differences}
Our counties demonstrated an average elevation of $1.129$ standard deviations in intra-county assessments of anxiety after COVID-19 incidence and a discontinuity slope of $-.21$ standard deviations after their first COVID-19 death.
This establishes a strong national trend consistent with other reviews \citep{xiong2020impact} and as we found in \autoref{fig:rdd_spaghetti_covid_case}. 
Before the incidence occurred, these same counties were lowering their levels of anxiety towards .02 standard deviations above their mean at a rate of -.38 standard deviations per week.
However, this trend belies high individual variance ($\sigma=1.032$) in anxiety shifts between counties (See \autoref{sec:idiosyncracies} for more details). 
As such, we find value in deciphering how a particular county will respond to a major event is a predictable phenomenon.

\subsection{Explaining Community-Idiosyncratic Effects}
To investigate the extent to which inherent differences between communities influence the predictive abilities of our estimates, we bin our counties by socio-economic status (SES) based on education and income into low (44 counties), medium (103 counties), and high (215 counties) SES.
For the most socioeconomically advantaged counties, we find that they express the highest increases in anxiety (an increase of $1.364 \pm .664$ deviations for high status, with $.923\pm .61$ and $.654\pm .757$ for medium and low status). 



Likewise, we separate our counties directly into tertiles based on urbanicity, which is calculated as the natural log of population divided by the county's surface area in square miles.
Similarly to socioeconomic status, we find that more urban communities experienced greater spikes in their anxiety levels (an increase of 1.62 deviations for high urbanicity, with 1.317 and .97 for medium and low).



Interestingly, for both status and urbanicity, we see that more rural and less well-off counties had a higher variance in how they responded than higher status and more urban ones.
This disparate impact of events on anxiety also brings up the concern that models may likewise display disparate performance. 
Indeed, when we examined the error of our models for high SES and low SES populations, they performed best on the low SES counties ($r = .84$) and worst on the high SES counties ($r=.71$), despite high SES being the most represented in the training set (See \autoref{sec:ablations} for plots). 
This fits with social scientific theory suggesting areas with low SES are focused on more fundamental needs (food, security, shelter)~\citep{maslow1943theory} and thus may have more predictable effects of events while high SES are more heterogeneous in terms of needs and resilience, and effects of events~\citep{oleson2004exploring}. 

Thus, results should be understood as having a wider bound on high-income urban communities, and the utilization of discontinuity forecasts within high SES communities should be treated with caution. 
Future work could explore more deeply the interplay between the specific identifiers of a community and its responses to a variety of major event types.

\begin{table*}[!t]
\centering
\begin{tabular}{lcccc} \toprule
 & \multicolumn{2}{c}{\textbf{Discontinuity}} & \multicolumn{2}{c}{\textbf{Discontinuity Slope}} \\ \cmidrule(lr){2-3}\cmidrule(lr){4-5} 
 & MSE ($\downarrow$) & r ($\uparrow$) & MSE ($\downarrow$) & r ($\uparrow$) \\ \midrule
Extra Trees                 & \textbf{.521} & \textbf{.760}$^\dagger$ & .031 & .861$^\dagger$ \\
Extra Trees (+exog) (+cov)  & .562 & .736 & .056 & .783 \\ 
Transformer                 & .652 & .678$^\dagger$ & .069 & .672$^\dagger$ \\
Transformer (+exog) (+cov)  & .526 & .753$^*$ & .039 & .840 \\ 
GRU                 & .628 & .713$^\dagger$ & .035 & .860$^{**}$ \\
GRU (+exog) (+cov)  & .521 & .755 & \textbf{.031} & \textbf{.868} \\ \bottomrule
\end{tabular}
\caption{Evaluating the use of sequential models. Significance testing was done on whether the added features improved the model over the base features.}
\label{tab:model_seq}
\end{table*}

\section{Related Work}
\paragraph{Machine Learning applied to Quasi-Experimental Designs}
Scientific research is founded upon the ability to establish causal relationships between causes and effects.
However, the use of purely data-driven statistical models, while impressive from a predictive or a correlation lens, is not sufficient to make causal claims, especially in the healthcare space~\citep{prosperi2020causal} or where predictions must be actionable~\citep{mothilal2020explaining}.
Natural Language Processing (NLP) and machine learning have largely concerned themselves with strictly predictive tasks but have recently begun to integrate causal inference and interpretations~\citep{verma2020counterfactual, feder2022causal}. 
In this vein, we have seen works that automatically identify inflection points around which to conduct RDDs, a task that typically requires rigorous human domain knowledge, using nonparametric structural changes, anomaly detection, and tree-based models~\citep{porter2015regression,herlands2018automated,liu2023automated}.
Relevant to this work are best practices for using textual data to generate outcomes, which recommend handling confounders through the use of held-out or disjoint sets of text for creating measures before applying those measures to identify causal effects~\citep{egami2022how, feder2022causal}.

\paragraph{Counterfactual Regression Over Time}
Past research explored the use of RMSNs~\citep{lim2018forecasting}, RNNs~\citep{bicaestimating}, and G-computation~\citep{li2020g} to conduct counterfactual regression over time, or forecasting a patient's treatment effects given the patient's history and outcomes. 
More recent works have explored the use of Transformers~\citep{melnychuk2022causal} and GRUs~\citep{NEURIPS2024_02cef2ae} to improve upon counterfactual regression over time.
The latter work found that RNNs optimized to generate representations most similar to the input space were capable of outperforming Transformers.
We contrast our work against counterfactual forecasting or calculating counterfactual outcomes~\citep{verma2024counterfactual} since we are concerned with separating idiosyncratic responses from national discontinuities in mental health.
In turn, this allows us to learn more about the discontinuity-causing event and shift our focus to community health in place of individual health.

\paragraph{Using Natural Language To Differentiate Group Mental Health}
Prior language-based works have found that there are differentiated responses to national events, such as the Murder of George Floyd, among demographics~\citep{eichstaedt2021emotional}. 
Other works have also identified innate differences in the psychological well-being of geographies~\citep{sturm2003geographic,jaidka2020estimating}. 
Among other psychological changes, we also see behavioral changes in the language use of social media users after major disasters~\citep{cohn2004linguistic, lin2014ripple}.
These methods present potential improvements over survey-based methods that can introduce self-report biases of participants and survey design~\citep{phillips1972some, bethlehem2010selection} because they readily distill the inherent context contained within language that could influence a population's mental health.
However, none of these human-centered approaches were capable of examining time-series effects for communities in parallel. 
Other more relevant works have used regression discontinuity designs within single geographical areas focused on how specific locales, e.g. spaces for public gathering, affected the expressed mental health of New Yorkers during the COVID-19 pandemic~\citep{jung2025changing}. 

\section{Discussion}

\subsection{Limitations}

Our work should be considered in light of language-based assessments (LBAs) and Longitudinal Regression Discontinuity Designs (LRDDs) being imperfect methods.
A limitation of language-based assessments is that to assess a community, a sizable amount of data must be collected, which is especially challenging for studying low-population communities. 
In the case of this work, it impacted our ability to study many of the more rural counties in the US that either have small populations or are not sufficiently active on social media platforms.  
We specifically utilized LBAs from the ``high-reliability setting'' (at least 200 people per county per week) which limited us to 388 counties, of which 361 had usable anxiety/depression timelines. Furthermore, while there was variance in the timing of COVID-19 events observed, it was limited to a 10-week period.
Note that LBAs cannot directly measure actual diagnoses of depression and anxiety, since differential diagnosis must be assessed in a clinical and personal setting. 
These assessments have been found to converge with community-scale self-reports and are nevertheless valuable for public health research.

Assertions of causality are always difficult to make in the absence of a fully randomized experiment. 
Quasi-experimental designs cannot be treated as a drop-in replacement for experimental designs or randomized control trials. 
However, in cases of population health, it is not feasible, or even ethical, to withhold treatment from random communities.
While estimating discontinuity bears similarity of an inference task it does not directly enable uncertainty quantification.
The specific model presented in this work is unlikely to generalize to timelines for other event types and LBMHA data for the relevant event would need to used in the training of relevant models.
If, for example, the new event chosen was a rare event like the opening of a major manufacturing center in a community there are risks of data sparsity which would affect a learned models ability to predict outcomes for communities in the future.
For works that extend the ideas herein to causal effects brought on by social categories, special care must be taken when manipulating categories such as race and gender~\citep{kasirzadeh2021use}.
In this work, we hope to make clear the value of an alternative research design that applies to the study of dense language-based assessments in time.
Other methods for conducting causal inferences, such as Difference in Differences (DiD) or Instrumental Variables (IV), are also suitable for this task, but we ultimately found that they do not offer sufficient interpretability. 
DiD has the additional challenge of generating and validating high-quality matches for calculating propensity scores (See \autoref{sec:diff_in_diff} for more on DiD estimation and matching techniques).

\subsection{Ethical Considerations}
Given that we work with data at the aggregate, non-individualized level, we release all community-level results and variables associated with this work publicly. 
All analyses were conducted over public, pre-existing data. 
Still, the methods in this work present potential risks if misused to conduct targeted political marketing, or violent actions against communities predicted to be more susceptible, based on response patterns observed in similar events. 
Such applications could lead to exploitation, manipulation, or harm, contradicting the intended purpose of this research.
However, we weigh these concerns against the positive impact of leveraging the LRDD paradigm with language assessments to better understand population health and enhance emergency response. 
Specifically, this line of work can lead to technologies that aid in the deployment of professionals and resources in anticipation of geographic regions most affected by an event.
For instance, it can facilitate the preparation and deployment of counselors trained to address anxiety in counties where heightened anxious reactions are anticipated, ensuring more effective and equitable crisis interventions.

\subsection{Conclusion}
Depressive and anxious disorders are among the top 20 leading causes of loss of disability-adjusted life year worldwide~\citep{vollset2024burden} and are highly costly illnesses to care for and to treat~\citep{konnopka2020economic}. Mental health is highly sensitive to changes in social context, such as the onset of a recession~\citep{katikireddi2012trends}, or the invasion of new infectious diseases~\citep{boden2021addressing}. 
Having a deeper longitudinal causal explanation of how critical events affect health empowers public health researchers to make firmer claims about the kinds of support that different communities require.
These results also validate previous temporal results found through surveys tracking the negative effects of COVID-19 brought to community anxiety~\citep{gallup2021covid}.
By utilizing language-based health assessments, we can take the next step toward applying quasi-experimental research designs more widely. These designs enable us to more closely and automatically generate causal inferences of how health events, interventions, and treatments affect mental health within populations. 

This study introduced the task of \task, used as a principled approach for assessing the community-specific effects of major events. 
We demonstrate the capacity of supervised learning techniques to predict community reactions through a combination of endogenous factors, dynamic covariates, and exogenous variables, offering a robust framework for anticipatory modeling. 
The error analysis highlights the extent to which sociodemographic differences and urbanicity contribute to observed variations, underscoring the importance of contextual factors in predictive modeling. 
Together, these findings enhance the methodological toolkit for computational social sciences, offering actionable insights into community resilience and the dynamics of external disruptions.
Finally, by evaluating the relationship between COVID-19 incidence and county-level depression and anxiety, we provide empirical evidence of the pandemic's psychological impact at a localized scale. 
Discontinuity forecasting expands the scope of predictive modeling beyond conventional approaches, offering a novel framework for estimating the distinct and localized effects of potential future or hypothetical events on specific communities.

\bibliography{custom}


\newpage 
\appendix

\section{Examining Idiosyncrasies of Communities}
\label{sec:idiosyncracies}

\begin{figure}[!t]
    \centering
    \includegraphics[width=.95\linewidth]{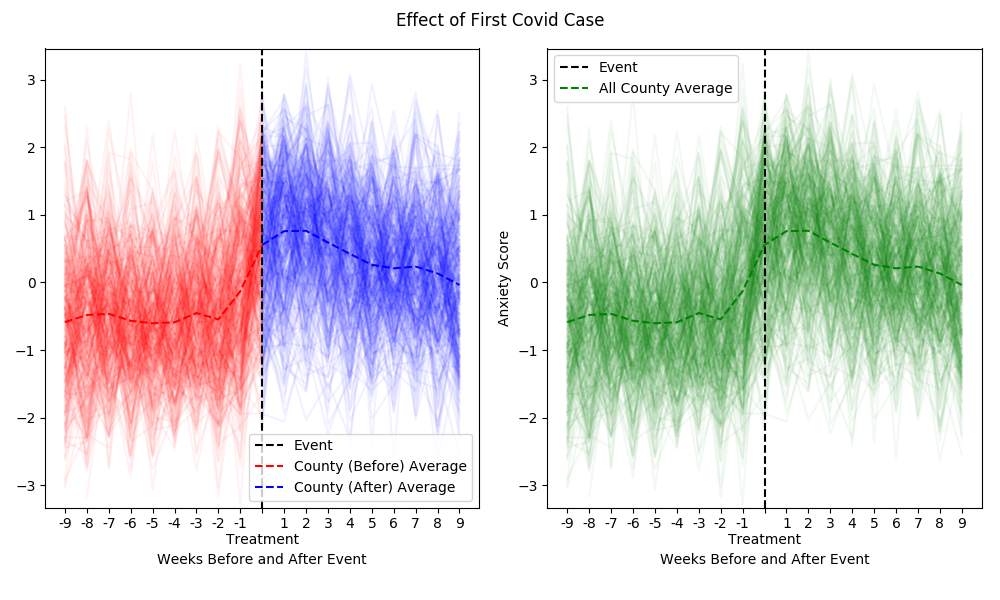}
    \caption{Individual lines show unique county anxiety assessment trends. To the right in green is the same data, but without distinguishing a pre- and post-event period.}
    \label{fig:covid_case_spaghettis}
\end{figure}

Despite strong national trends in changes to anxiety as a result of COVID-19 onset, there is a meaningful range of responses across different US counties. In \autoref{tab:stats_outcomes} we show descriptive statistics for the discontinuity shift (Discontinuity), slope (Discontinuity slope), observed anxiety at the time of the event (Anx at $t=0$), and the week after (Anx at $t=1$), and the regression parameters ($\beta_0$ and $\beta_1$ Before) for the period from $-T \leq t \leq 1$. \autoref{fig:covid_case_spaghettis} makes this point visually clear by depicting the range of anxiety scores present in each county leading up to their first incidence of COVID-19, alongside the national story. 


\begin{table}[!h]
    \centering
    \begin{tabular}{lrrr}
        \toprule
        Statistic & Mean & Std Dev & Median \\
        \midrule
        Discontinuity & 1.129 & 1.032 & 1.197 \\
        Discontinuity Slope & -.096 & .289 & -.135 \\
        Anx at $t=0$ & .558 & 1.038 & .634 \\
        Anx at $t=1$ & .759 & 1.012 & .887 \\
        $\beta_0$ Before & .017 & .194 & .028 \\
        $\beta_1$ Before & -.375 & .882 & -.386 \\
        \bottomrule \\
    \end{tabular}
    \caption{Descriptive statistics for anxiety before the 361 counties' first COVID-19 hospitalization.}
    \label{tab:stats_outcomes}
\end{table}

\section{Ablations}
\label{sec:ablations}

\subsection{Buffering Events}
\begin{figure}[!t]
    \centering
    \begin{subfigure}[t]{.495\linewidth}
        \centering
        \includegraphics[width=\linewidth]{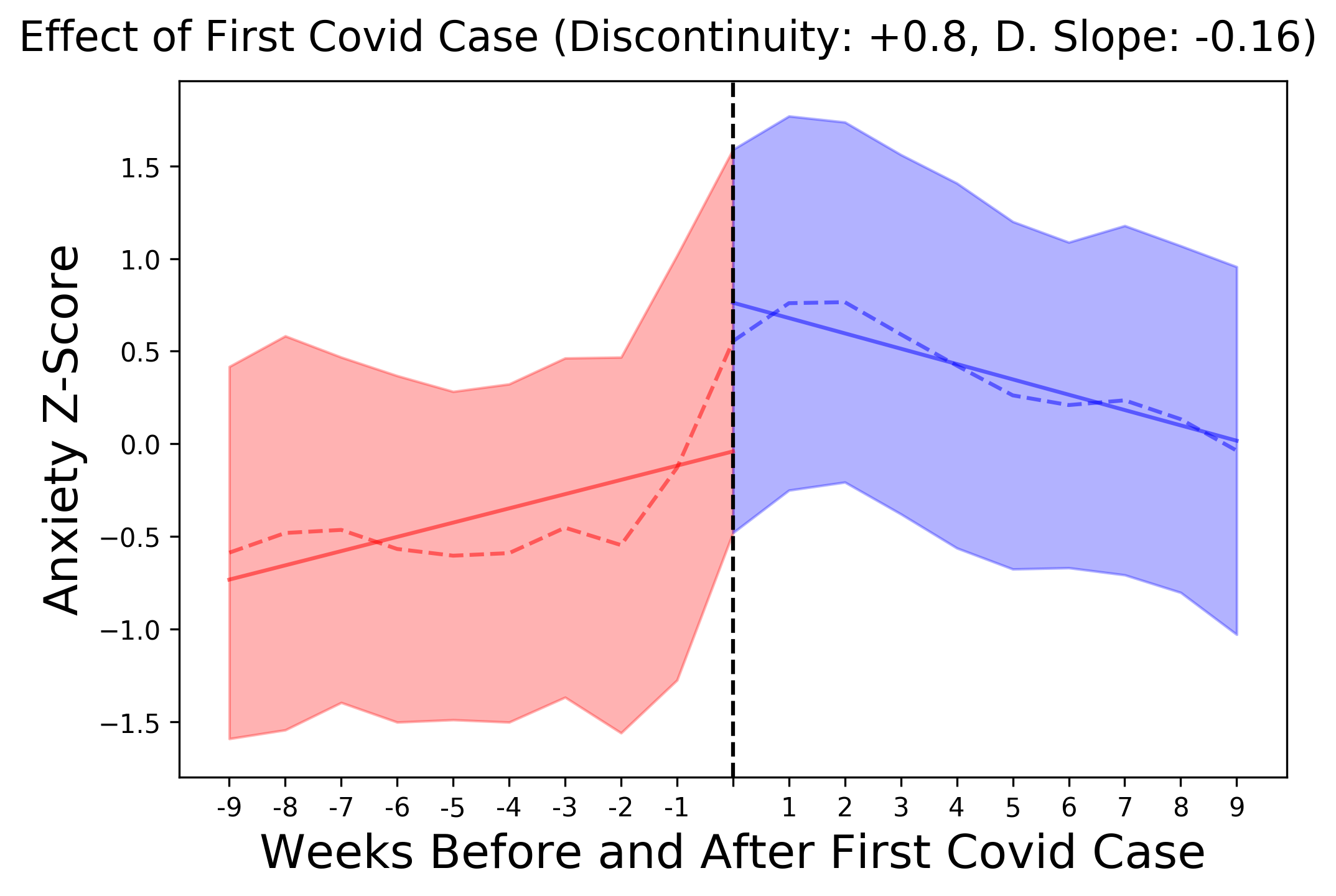}
    \end{subfigure}
    \begin{subfigure}[t]{.495\linewidth}
        \centering
        \includegraphics[width=\linewidth]{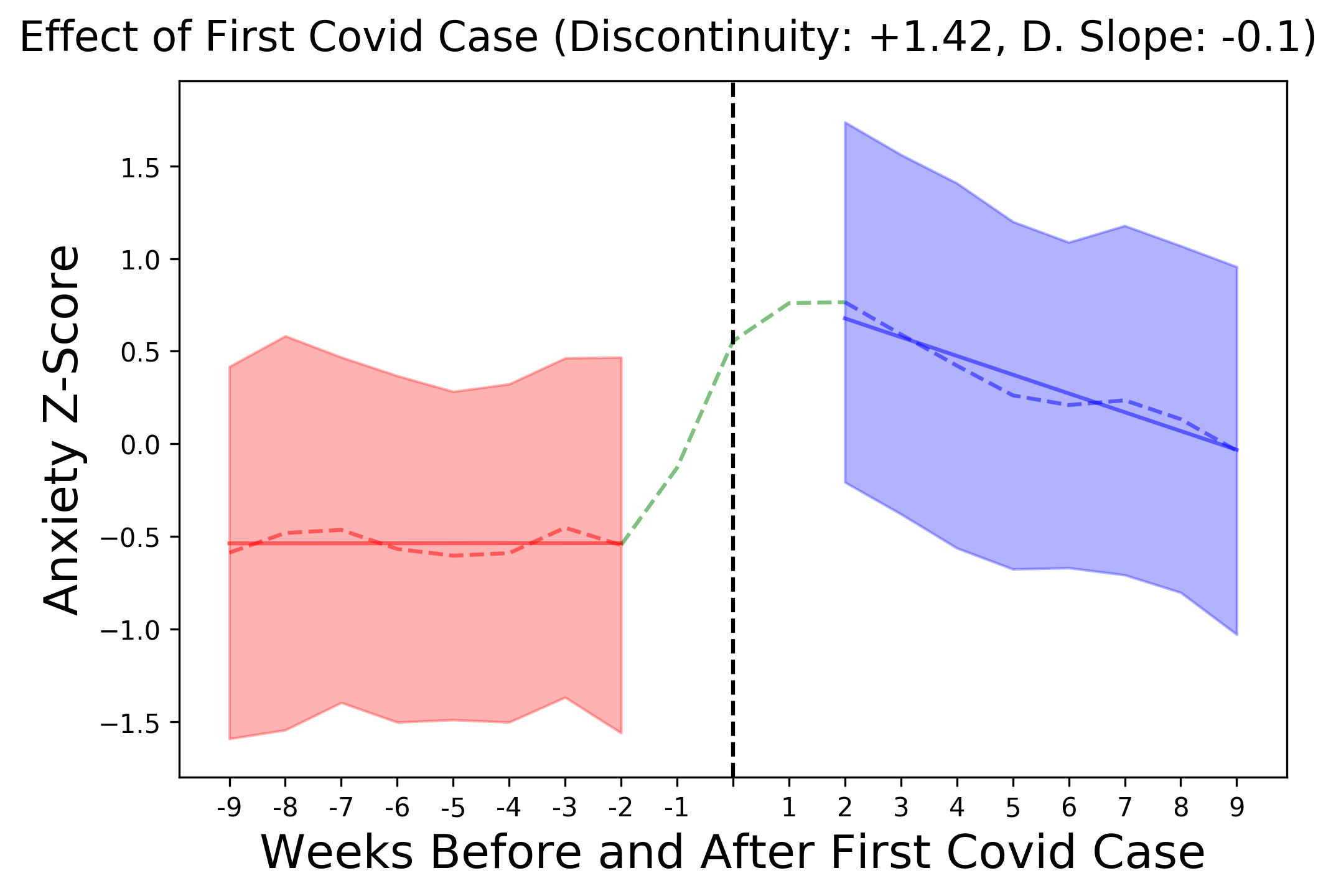}
    \end{subfigure}
    \caption{Anxiety before and after the first case of COVID-19 with no buffer (a) and a buffer size of 2 (b) applied}
    \label{fig:covid_case_buffering}
\end{figure}

To account for the effect of general increases in anxiety before a major event and time for the event news to be disseminated within a given community, we used a buffer of 1 week around the event. To confirm that the effect being studied was still present and not highly sensitive to the buffer size we demonstrate the change from using a buffer of 0 or 2 weeks in \autoref{fig:covid_case_buffering}. The authors recommend the use of at least a 1 week buffer on events on televised events, it is likely that alternative buffers will be required for different event types.

\subsection{General Trends by Socio-Economics}

\begin{figure}[!t]
    \centering
    \includegraphics[width=.6\linewidth]{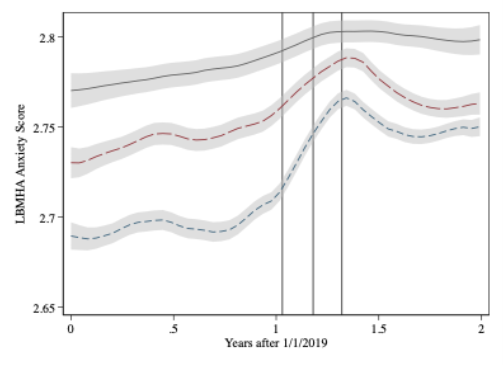}
    \caption{Non-linear trends in Anxiety throughout 2019-2020. Results are stratified by counties into socio-economic tertiles and range from low (solid black line) through average (maroon long dashed line) to high SES (teal dashed line). 95\% confidence intervals are provided in gray. Vertical black lines identify the commencement of the US COVID-19 epidemic (first line from the left) and the start/end of the initial US shutdowns/work-from-home orders.}
    \label{fig:anxiety_trends_by_ses}
\end{figure}

We visualized changes in LBMHA anxiety and depression over time before and through the COVID-19 pandemic, stratified by socioeconomic status (SES) \autoref{fig:anxiety_trends_by_ses}. In this analysis, you can see that there was a shift upward in anxiety that began before the first confirmed COVID-19 case in the USA (marked by the first vertical black line) and continued uninterrupted through the shutdowns (demarcated on the graph by the second and third black lines at March 16th and July 1st, 2020) until the summer of 2020, whereupon they ebbed somewhat and ultimately stabilized at much higher levels across all SES groups. Critically, those in the lowest SES group (solid black horizontal lines) started at the highest levels and then showed relative stability throughout the pandemic.

\begin{figure}[!t]
    \centering
    \begin{subfigure}[t]{.495\linewidth}
        \includegraphics[width=\linewidth]{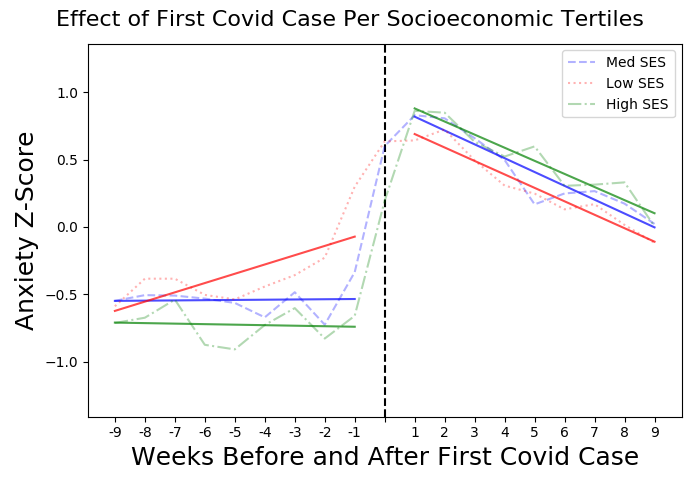}
        \caption{}
        \label{fig:covid_case_ses}
    \end{subfigure}
    \begin{subfigure}[t]{.495\linewidth}
        \includegraphics[width=\linewidth]{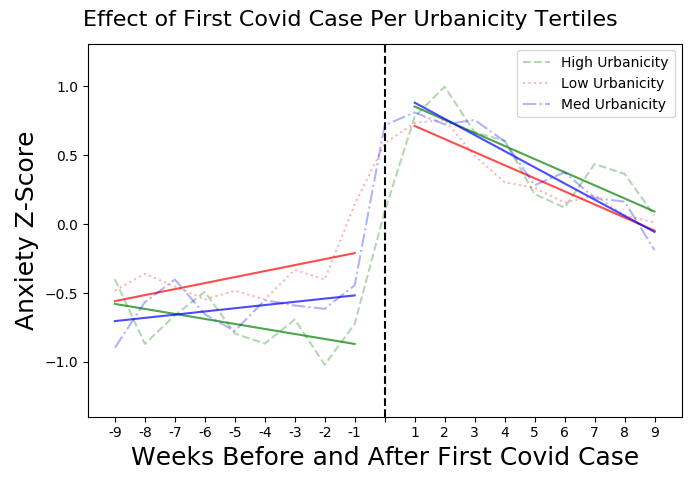}
        \caption{}
        \label{fig:covid_case_urbanicity}
    \end{subfigure}
    \caption{Discontinuities in Anxiety by Socio-Economic Status (a) and Urbanicity (b) before and after the first case of COVID-19 reaches a county.}
    \label{fig:covid_case_error_analysis}
\end{figure}

\autoref{fig:covid_case_ses} and \autoref{fig:covid_case_urbanicity} show the disparate impact of COVID cases across different socio-economic statuses and urbanicity.

\section{Model Hyperparameters}
\label{sec:hyperparameters}

The computational load to run our models was rather modest, and a complete recreation of our results could be completed on a CPU (Intel Xeon E5-2630) in under 3 hours. Optionally, using a single GPU (NVIDIA Titan Xp) it is possible to do the same in under 30 minutes. For training sequential contextual, we used Optuna with a TPESampler and the HyperbandPruner.
All experiments are conducted with the same random seed or state, which was never altered. 
A complete enumeration of the hyperparameters used is in \autoref{tab:model_configurations}.

\begin{table}[!t]
\centering
\begin{tabular}{lcccc} \toprule
 & \multicolumn{2}{c}{\textbf{Discontinuity}} & \multicolumn{2}{c}{\textbf{Discontinuity Slope}} \\ \cmidrule(lr){2-3}\cmidrule(lr){4-5} 
 & MSE ($\downarrow$) & r ($\uparrow$) & MSE ($\downarrow$) & r ($\uparrow$) \\ \midrule
\multicolumn{5}{l}{\textbf{with endogenous:}}\\
\midrule
RC (2)      & .613 & .713$^\dagger$ & .032 & .866$^\dagger$              \\
P (9)       & .618 & .709$^\dagger$ & \textbf{.031} & .865$^\dagger$              \\
P + RC (11) & .620 & .709$^\dagger$ & .031 & .865$^\dagger$ \\ \midrule
\multicolumn{5}{l}{\textbf{with exogenous and covariates:}}\\
\midrule
exog (1024)            & 1.147  & .257$^*$ & .118 & .223$^*$ \\
exog + RC (1026)       & \textbf{.600}  & \textbf{.716}$^\dagger$ & .032 & .863$^\dagger$ \\
exog + P (1033)        & .607  & .713$^\dagger$ & .031 & .866$^\dagger$ \\
exog + P + RC (1035)   & .608  & .713$^\dagger$ & .031 & .866$^\dagger$ \\
cov + P + RC  (361)   & .604  & .714$^\dagger$ & .033 & .867$^\dagger$ \\ 
cov + exog + P + RC   (1046)   & .602  & .715$^\dagger$ & .033 & \textbf{.868}$^\dagger$ \\ \bottomrule \\
\end{tabular}
\caption{Evaluation Comparison of feature sets.  
$P$ is the set of points preceding the event being measured from $t=-9$ to $t=-1$, $RC$ is a tuple containing the $\beta_0$ and $\beta_1$ coefficients of the regression line for the before period. $exog$ represents the use of RoBERTa-Large county embeddings. $cov$ represents the dynamic covariates, i.e. language-based depression.
Here we report mean squared errors and Pearson correlations for predicting discontinuities and discontinuity slopes of Anxiety scores before a county has its first case of COVID-19. All models were trained using Ridge Regression with an alpha of $1.0$.\\
Significance testing of whether the features improve performance over the baseline.}
\label{tab:picking_features}
\end{table}

\begin{table}[ht]
    \centering
    \begin{tabular}{p{.3\linewidth}  p{.6\linewidth}}
        \toprule
        Model & Hyperparameters \\
        \midrule
        Ridge (+exog+cov)  & Alpha = 1.0 (10.0) \\
        kNN (+exog+cov)  & Neighbors = 5 (5), Metric = Euclidean Distance (Euclidean Distance) \\
        Random Forest (+exog+cov)  & Estimators = 500 (1000), Max Depth = None (None) \\
        Extra Trees (+exog+cov)  & Estimators = 500 (500), Max Depth = 10 (None) \\ 
        XGBoost (+exog+cov)  & Estimators = 500 (1000), Max Depth = None (None) \\ \midrule
        FFN  & Epochs = 150, LR = .005, Layers = 2, Layer Size = 2   \\
        GRU (+exog+cov)  & Epochs = 50 (75), LR = .005 (.0005), Layers = 4 (2), Layer Size = 100 (500), Dropout = .1 (.5) \\
        Transformer (+exog+cov) & Epochs = 200 (100), LR = .005 (.0001), Layers = 2 (2), Layer Size = 10 (500), Dropout = .5 (.1), Heads = 1 \\
        \bottomrule \\
    \end{tabular}
    \caption{All sequential models (FFN, GRU, Transformers) were trained with Hyperband pruning via Optuna, Adam optimization, a batch size of 64, and a 60-20-20 Train-Val-Test split. All other models were trained using GridSearchCV with 5 Folds for Cross Validation.}
    \label{tab:model_configurations}
\end{table}

\section{Baselines}
\label{sec:baselines}

\subsection{Mean Baseline}
For this baseline we always predict the mean intercept ($\Bar{\delta_0}$) and slope ($\Bar{\delta_1}$) discontinuity of the training set.
We consider this baseline as a reasonable floor for performance based on the occurrence of the event with no notion of community differences. Since we find that this baseline is outperformed by simple models we conclude that event incidence is not sufficient for Discontinuity Forecasting.

\subsection{No Change Baseline}
For this baseline we always predict that the linear fit parameters ($\hat\beta_0^{(\text{before})}$,$\hat\beta_1^{(\text{before})}$) describing the points before $t=0$ will hold. This effectively means always predicting that $\delta_0=0$ and $\delta_1=0$. 
This tests whether the null hypothesis that the event has no effect, which we find to be false.

\subsection{Forecasting Baseline}
For this baseline we fit an ARIMA model on the data points before the event from $t=-T$ to  $t=-1$. Then we predict all future points up to $t=T$, then infer the slope and intercept and subtract the prior estimates to find the deltas. All ARIMA order parameters were chosen automatically using auto-ARIMA applied locally to each timeline. We find that simple forecasting is insufficient for Discontinuity Forecasting.

\section{Difference in Differences}
\label{sec:diff_in_diff}

\subsection{Matching}
To match counties for creating synthetic counterfactual counties, we adopted concepts from propensity scores.. In our implementation, each candidate matching county is represented as a contextual 9-length vector relative to all other target counties. 
The first three dimensions contain a PCA dimension reduction of sociodemographic factors for each county.
The next two dimensions contain the latitude and longitude of the center of the county. 
These vectors are then compressed via PCA into 3 components. 
The following dimension is a binary contextual factor that measures if the candidate county is physically adjacent to the target county.
The final three dimensions contain the treatment score, triplicated, for the candidate county before the target county had its intervention, i.e. did the candidate county have a similar outcome score as the target county before the studied intervention occurred?
Counties can now be matched with their most similar county pairs based on the Euclidean distance of their vector representation to that of all other counties.

\subsection{Diff. in Diff. Estimation}
The calculation of difference-in-difference is done by measuring the difference between the expected change in the running variable minus the observed change. 

Counties are labeled as $y_{i,t}$ where i is the county being studied (0 for the average matched county, 1 for the target county) and t is the time step of the study (0 before intervention, 1 after intervention)
To measure the counterfactual of the target county after the intervention ($\tilde{y}_{1,1}$), we add the change of the average matched county to the target county.
\begin{equation}
    \tilde{y}_{1,1} = y_{1,0} + (avg(y_{0,1}) - avg(y_{0,0})) \\
\end{equation}

Then the calculation of the difference in differences ($DiD$) is expressed as the difference between how the target county was observed to behave minus how it was expected to behave:
\begin{equation}
    DiD = y_{1,1} - \tilde{y}_{1,1}
\end{equation}

\begin{figure}[!t]
    \centering
    \includegraphics[width=.8\linewidth]{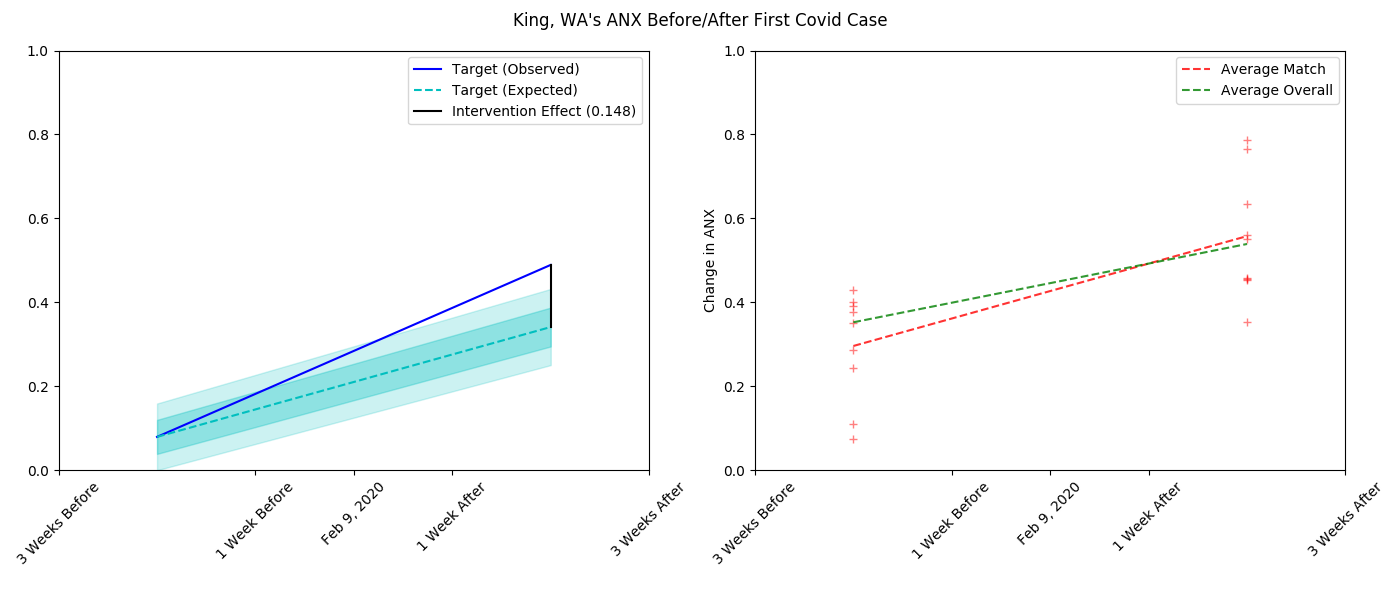}
    \caption{Example of the diff-in-diff pattern of quasi-experimental design. Here we see the example change in the language-based assessment of Anxiety for King County in Washington state.}
    \label{case_studies}
\end{figure}

\end{document}